\documentclass{article}

\usepackage[final,nonatbib]{neurips_2018}

\usepackage{tikz}
\usetikzlibrary{shapes.callouts}

\usepackage{mathpazo}
\usepackage{relsize}
\usepackage[T1]{fontenc}
\usepackage[scaled=0.90]{inconsolata}
\usepackage[latin1]{inputenc}
\usepackage[english]{babel}

\usepackage{epsfig}
\usepackage{graphicx}
\usepackage{wrapfig}
\usepackage[belowskip=0pt,aboveskip=3pt,font=small]{caption}
\usepackage[belowskip=0pt,aboveskip=0pt,font=small]{subcaption}
\usepackage{amsmath, amsthm, amssymb}
\usepackage{textcomp}
\usepackage{stmaryrd}
\usepackage{upgreek}
\usepackage{bm}
\usepackage{cases}
\usepackage{mathtools}
\usepackage{arydshln}
\usepackage{multirow}

\usepackage{cite}
\definecolor{bluebell}{RGB}{52,31,151}
\definecolor{amour}{RGB}{238,82,83}
\usepackage[pagebackref=true,breaklinks=true,letterpaper=true,colorlinks,bookmarks=false,citecolor=amour,linkcolor=bluebell]{hyperref}

\usepackage{algorithm}
\usepackage{algpseudocode}

\usepackage{multirow}
\usepackage{rotating}
\usepackage{booktabs}

\usepackage{enumitem}
\usepackage[olditem,oldenum]{paralist}

\usepackage{alltt}
\usepackage{listings}

\usepackage{url}
\usepackage{xspace}
\usepackage{comment}
\usepackage{color}
\usepackage{xcolor}
\usepackage{afterpage}
\usepackage{pdfpages}
\usepackage{framed}
\usepackage{fancybox}
\usepackage{cuted}
\usepackage{caption}

\usepackage{mysymbols}

\newcommand{\vd}{VisDial\xspace}

\newcommand{\myquote}[1]{\emph{`#1'}}

\newcommand{\lf}{LF\xspace}

\newcommand{\hre}{HRE\xspace}
\newcommand{\mn}{MN\xspace}

\newcommand{\reftab}[1]{Tab.~\ref{#1}}

\newlength{\bibitemsep}
\newlength{\bibparskip}
\let\oldthebibliography\thebibliography
\renewcommand\thebibliography[1]{%
  \oldthebibliography{#1}%
  \setlength{\parskip}{\bibitemsep}%
  \setlength{\itemsep}{\bibparskip}%
}

\title{{\Large Response to ``Visual Dialogue without Vision or Dialogue"~\cite{massiceti_nipsw18}}}

\author{
    Abhishek Das \hspace{0.5pc}
    Devi Parikh \hspace{0.5pc}
    Dhruv Batra \\[0.05in]
    {\small Georgia Institute of Technology} \\[0.05in]
    {\tt\small \{abhshkdz, parikh, dbatra\}@gatech.edu}}

\begin{document}

\maketitle

\begin{abstract}
    In a recent workshop paper, Massiceti~\etal~\cite{massiceti_nipsw18} presented
    a baseline model and subsequent critique of Visual Dialog~\cite{das_cvpr17}
    that raises what we believe to be unfounded concerns about the dataset and evaluation.
    This article intends to rebut the critique and clarify potential confusions for
    practitioners and future participants in the Visual Dialog challenge.
\end{abstract}

\section{Introduction}.
\vspace{-10pt}

\begin{figure}[h]
  \centering
  \includegraphics[width=\linewidth]{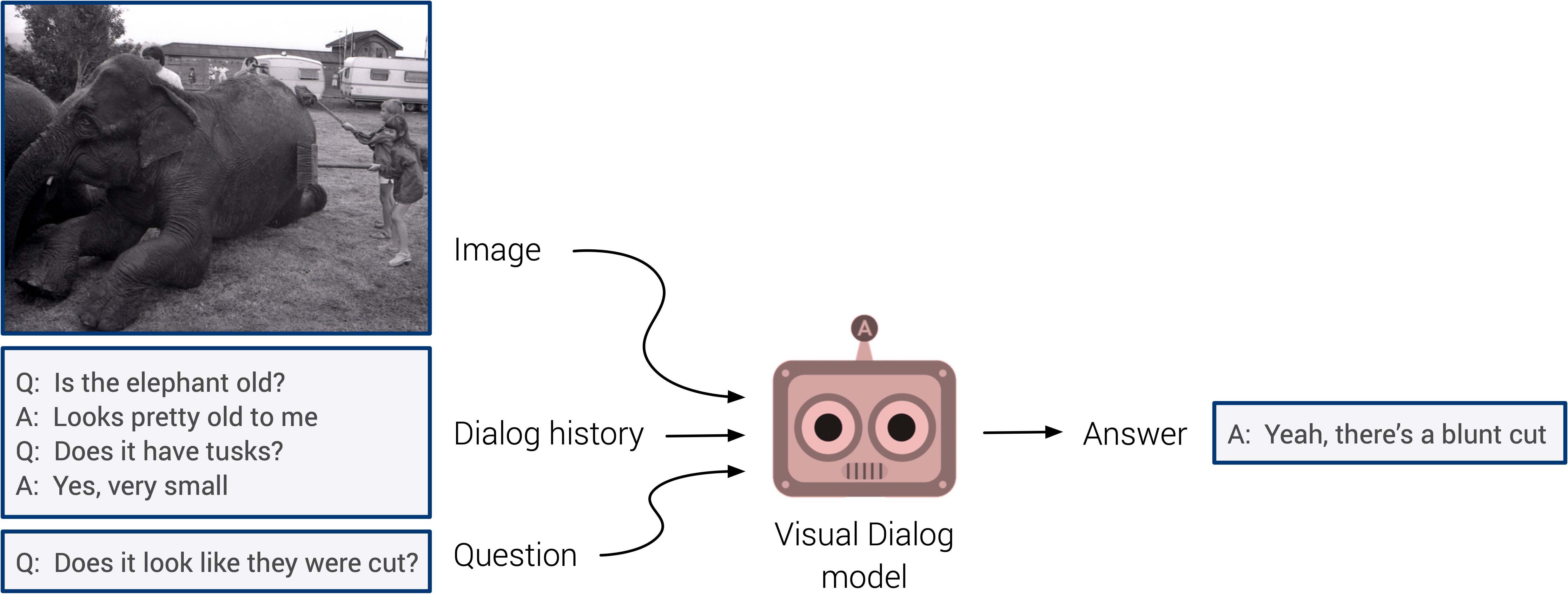}
  \caption{Visual Dialog task: given an image, dialog history, and follow-up question, predict the answer.}
  \label{fig:task}
\end{figure}

\textbf{Task}.
The goal of Visual Dialog is to develop conversation agents that can talk about images.
Towards this end, in previous work~\cite{das_cvpr17}, we proposed a task --
given an image, dialog history, and follow-up question, predict a free-form natural language answer
to the question (\figref{fig:task}) -- and a large-scale dataset\footnote{\href{https://visualdialog.org/data}{\texttt{visualdialog.org/data}}},
evaluation metrics and
server\footnote{\href{https://evalai.cloudcv.org/web/challenges/challenge-page/103/overview}{\texttt{evalai.cloudcv.org/web/challenges/challenge-page/103/overview}}},
and baseline models\footnote{\href{https://github.com/batra-mlp-lab/visdial}{\texttt{github.com/batra-mlp-lab/visdial}}}
for this task.

\textbf{Key challenge}. A fundamental challenge in dialog systems is automatic evaluation of long free-form answers
since existing metrics such as BLEU, METEOR, and ROUGE are known to correlate poorly with human
judgement~\cite{liu_emnlp16}. Thus,
as proposed in our initial paper~\cite{das_cvpr17}, to evaluate Visual Dialog,
models are provided a list of $100$ candidate
answers for each question -- consisting of the ground-truth answer from the dataset mixed with
nearest neighbors, popular, and random answers -- and evaluated on how well they rank the
ground-truth answer on retrieval metrics such as
mean reciprocal rank (MRR), recall (R@{1, 5, 10}), and mean rank.

As we describe in our paper~\cite{das_cvpr17}, these candidate answers for each
question are programmatically curated from other answers in the dataset and not
human-generated, and so, some candidate answers may be semantically identical (e.g. `yeah' and `yes').
Thus, more recently, we conducted new human studies -- asking four human subjects to
annotate whether each of the $100$ candidate answers is correct or not
for all questions in the VisDial test split. For evaluation, we report the normalized discounted
cumulative gain (NDCG) over the top $K$ ranked options, where $K$ is the number of answers marked
as correct by at least one annotator. For this computation, we consider the relevance of an
answer to be the fraction of annotators that marked it as correct. This was the primary evaluation
criterion for the $1$st
Visual Dialog Challenge\footnote{\href{https://visualdialog.org/challenge/2018\#evaluation}{\texttt{visualdialog.org/challenge/2018\#evaluation}}}.

As described in~\cite{das_cvpr17}, there are two broad families of dialog models
(unfortunately with names that are overloaded in machine learning) -- `generative'
models (that produce a response word-by-word given some context
and are evaluated on the ranking of the likelihood scores they assign
to candidate answers), and `discriminative' models (that simply learn to rank a list of candidate answers
and cannot produce a new response).
This retrieval-based evaluation holds for both families.
Compatibility of the evaluation metric with
generative models is crucial, since they are more useful for real-world applications
where answer options are not available.

\section{Concern 1: Suitability of NDCG evaluation}.
\vspace{-10pt}

Massiceti~\etal~\cite{massiceti_nipsw18} note that
\myquote{the VisDial dataset was recently updated to version $1.0$, where the curators try to
    ameliorate some of the issues with the single-``ground-truth" answer approach. They incorporate a
    human-agreement scores for candidate answers, and introduce a modified evaluation which weighs
    the predicted rankings by these scores.
    However, in making this change, the primary evaluation for this data has
    now become an explicit classification task on the candidate answers -- requiring access,
    at train time, to all $100$ candidates for every question-image pair. For the stated goals
    of Visual Dialog, this change can be construed as unsuitable as it falls into the
    category of redefining the problem to match a potentially unsuitable evaluation measure --
    how can one get better ranks in the candidate-answer-ranking task.}

The claim that ``the primary evaluation for this data has now become an explicit
classification task on the candidate answers''
is incorrect and thus the conclusion drawn from it is inaccurate and confusing.
First, the task has not changed, only the evaluation metric (from MRR to NDCG).
The task did not and does not ``require access, at train time, to all $100$ candidates''.
Discriminative models use $100$ candidate answers at train time; generative models do not.
This was discussed in our initial paper~\cite{das_cvpr17} and continues to be true.

Perhaps what the authors~\cite{massiceti_nipsw18} are trying to say and express concern
for is -- this metric (NDCG) will favor one kind of model family over another.
This is possible and something we have given a lot of thought to.
Empirical findings from
the $1$st Visual Dialog
Challenge\footnote{\href{https://visualdialog.org/challenge/2018\#winners}{\texttt{visualdialog.org/challenge/2018\#winners}}}
indicate that these generative models perform comparably (or even better sometimes)
than discriminative models on the NDCG metric -- for example, $53.67$ \vs $49.58$
on VisDial v$1.0$ test-std for Memory Network
$+$ Attention with generative \vs discriminative decoding respectively.
Code and models available here: \href{https://github.com/batra-mlp-lab/visdial\#pretrained-models-1}{\texttt{https://github.com/batra-mlp-lab/visdial\#pretrained-models-1}}.
While this is still a potentially weak surrogate for human-in-the-loop evaluation
of Visual Dialog models, it is encouraging that there now seems to be an automatic evaluation criterion
on which generative models, which do not have access to candidate answers during training,
outperform discriminative models.
As we describe on \href{https://visualdialog.org/challenge/2018#faq}{\texttt{visualdialog.org}},
the reason why we chose a single track for the challenge was that in practice, the distinction between
the two model families can get blurry (e.g., non-parametric models that internally
maintain a large list of answer options), and the separation would be difficult to enforce.
Note that our choice of ranking for evaluation isn't an endorsement of either
approach (generative or discriminative).

\clearpage
\section{Concern 2: Comparison to proposed CCA baseline~\cite{massiceti_nipsw18}}.
\vspace{-10pt}

Massiceti~\etal~\cite{massiceti_nipsw18} proposed a simple CCA baseline
with two variants -- 1) question-only (ignoring image and dialog history),
2) question $+$ image (ignoring dialog history), which they show
outperforms state-of-the-art models on the mean rank metric.
They further note that
\myquote{an important takeaway from our analyses is that it is highly effective to begin
    exploration with the simplest possible tools one has at one's disposal. This is particularly apposite in
    the era of deep neural networks, where the prevailing attitude appears to be that it is preferable to
    start exploration with complicated methods that aren't well understood, as opposed to older, perhaps
    even less fashionable methods that have the benefit of being rigorously understood.}

We agree that simple and strong baselines are important,
and are pleasantly surprised to see that a CCA baseline performs so well on mean rank.
However, there are a few problems with this analysis.
First, the baseline proposed by Massiceti~\etal~\cite{massiceti_nipsw18} is not close to state-of-the-art --
the authors cherry-pick the mean rank metric and ignore trends on \emph{all other metrics} (see~\reftab{table:model_results}).
Second, it ignores that a similar finding has already been presented in the original
Visual Dialog paper~\cite{das_cvpr17}, that question-only and question $+$ image models perform close to but
slightly worse than full Q$+$I$+$H models. We recreate~\reftab{table:model_results} from~\cite{das_cvpr17}.
Third, the authors~\cite{massiceti_nipsw18} ignore that the CCA baselines
perform worse than not just state-of-the-art models, but also
these Q and Q$+$I ablations~\cite{das_cvpr17},
and comparable to answer prior and nearest neighbor (NN)
baselines~\cite{das_cvpr17} on MRR and R@k.
Finally, the results presented in~\cite{massiceti_nipsw18} are not directly comparable.
The proposed CCA baselines use Resnet-34~\cite{he_cvpr16} features and
FastText~\cite{bojanowski_tacl17} embeddings, while the baselines in~\cite{das_cvpr17}
use VGG-16~\cite{simonyan_iclr15} and learn word embeddings from scratch respectively.

\begin{table}[h]
{
    \small
    \setlength\tabcolsep{3.8pt}
    \centering
    \begin{tabular}{cccccccc}
    \toprule
    & \textbf{Model} & \textbf{NDCG} & \textbf{MRR} & \textbf{R@1} & \textbf{R@5} & \textbf{R@10} & \textbf{Mean Rank} \\
    \midrule
    \multirow{3}{*}{\rotatebox[origin=c]{90}{v$0.9$ \texttt{val}} $\begin{dcases} \\ \\ \\ \\ \\ \\ \\ \\ \\ \end{dcases}$}
    &Answer prior & - & 0.3735 & 23.55 & 48.52 & 53.23 & 26.50 \\[1.5pt]
    &NN-Q & - & 0.4570 & 35.93 & 54.07 & 60.26 & 18.93 \\[1.5pt]
    &NN-QI & - & 0.4274 & 33.13 & 50.83 & 58.69 & 19.62 \\[1.5pt]
    &\lf-Q-G & - & 0.5048 & 39.78 & 60.58 & 66.33 & 17.89 \\
    &\lf-QI-G & - & 0.5204 & 42.04 & 61.65 & 67.66 & 16.84 \\
    &\lf-QIH-G & - & 0.5199 & 41.83 & 61.78 & 67.59 & 17.07 \\
    &\hre-QIH-G & - & 0.5237 & \textbf{42.29} & 62.18 & 67.92 & 17.07 \\
    &\hre{}A-QIH-G & - & 0.5242 & 42.28 & 62.33 & 68.17 & 16.79 \\
    &\mn-QIH-G & - & \textbf{0.5259} & \textbf{42.29} & \textbf{62.85} & \textbf{68.88} & 17.06 \\
    \cdashline{2-8}
    &A-Q (Massiceti~\etal~\cite{massiceti_nipsw18}) & - & 0.3031 & 16.77 & 44.86 & 58.06 & \textbf{16.21} \\
    &A-QI (Massiceti~\etal~\cite{massiceti_nipsw18}) & - & 0.2427 & 12.17 & 35.38 & 50.57 & 18.29 \\
    \midrule
    \multirow{1}{*}{\rotatebox[origin=c]{90}{v$1.0$ \texttt{test-std}} $\begin{dcases} \\ \\ \\ \\ \end{dcases}$}
    &\lf-QIH-G & 0.5121 & 0.4568 & \textbf{35.08} & 55.92 & \textbf{64.02} & 18.81 \\
    &\hre-QIH-G & 0.5245 & 0.4561 & 34.78 & 56.18 & 63.72 & 18.78 \\
    &\mn-QIH-G & \textbf{0.5280} & \textbf{0.4580} & 35.05 & \textbf{56.35} & 63.92 & 19.31 \\
    \cdashline{2-8}
    &A-Q (Massiceti~\etal~\cite{massiceti_nipsw18}) & - & 0.2832 & 15.95 & 40.10 & 55.10 & \textbf{17.08} \\
    &A-QI (Massiceti~\etal~\cite{massiceti_nipsw18}) & - & 0.2393 & 12.73 & 33.05 & 48.68 & 19.24 \\
    \bottomrule
    \end{tabular}
    \vspace{5pt}
    \caption{Performance of methods on \vd v$0.9$ and v$1.0$, measured by
    normalized discounted cumulative gain (NDCG), mean reciprocal rank (MRR), recall@$k$ and mean rank.
    Higher is better for NDCG, MRR, and recall@k, while lower is better for mean rank.}
    \label{table:model_results}
}
\end{table}

\section{Conclusion}
To summarize:
\begin{compactitem}
\item In an attempt to make evaluation for Visual Dialog more reliable,
    we have recently had multiple human subjects indicate whether each of the candidate answers for a question is
    correct, which is then used as the reference score while computing the NDCG metric.
    Massiceti~\etal\cite{massiceti_nipsw18} claim that this changes the Visual Dialog task to an ``explicit
    classification task on the candidate answers'' which is incorrect. The task remains the same as before~\cite{das_cvpr17},
    only the evaluation has changed.
\item Further, NDCG evaluation using dense annotations does not favor a particular
    family of Visual Dialog models (between discriminative and generative),
    as evidenced by findings from the $1$st Visual Dialog challenge noted on
    \href{https://visualdialog.org/challenge/2018\#faq}{\texttt{visualdialog.org}}.
\item While we welcome simple and strong baselines, the CCA baseline for Visual Dialog
    proposed by Massiceti~\etal~\cite{massiceti_nipsw18}
    is not close to state-of-the-art. The authors solely focus on one metric (mean rank) while ignoring
    all other metrics (MRR, R@k, NDCG) on which their approach is significantly worse, not just
    against state-of-the-art models, but also against ablations from~\cite{das_cvpr17} (see~\reftab{table:model_results}).
\item Finally, the VisDial dataset~\cite{das_cvpr17} and evaluation are not perfect unbiased testbeds.
    VisDial likely has many biases and trivial correlations models can pick up on,
    as has been previously observed in other unstructured (or loosely structured)
    real-world datasets~\cite{goyal_cvpr17}.
    Further, automatic evaluation of dialog is an open research problem,
    and our NDCG evaluation protocol for Visual Dialog is an attempt at
    making it more robust. Alternatively, evaluation with humans paired
    with dialog models, conversing for the human to be able to achieve a
    downstream goal (\eg understand their visual surroundings, book a flight ticket,~\etc)
    would perhaps be the truest form of evaluation, as has been explored in~\cite{chattopadhyay_hcomp17},
    although this is expensive. There is scope for improvement across all axes --
    task/dataset, evaluation, as well as methods.
\end{compactitem}

\begingroup
    {\small
        \setlength{\bibitemsep}{0.75\baselineskip plus .05\baselineskip minus .05\baselineskip}
        \bibliographystyle{ieee}
        \bibliography{strings,main}}
\endgroup

\end{document}